\documentclass[11pt]{article}

\usepackage[a4paper,margin=1in]{geometry}
\usepackage{microtype}

\usepackage{fontspec}

\usepackage{amsmath,amssymb,mathtools}
\usepackage{booktabs}
\usepackage{tabularx}
\usepackage{multirow}
\usepackage{array}
\usepackage{graphicx}
\graphicspath{{./}}
\usepackage{subcaption}
\usepackage{caption}
\usepackage{placeins}  
\usepackage{float}
\usepackage{tikz}
\usetikzlibrary{arrows.meta, positioning, calc, fit,
                shapes.geometric, shapes.misc, backgrounds,
                decorations.pathreplacing}

\usepackage{algorithm}
\usepackage{algpseudocode}

\usepackage[inline]{enumitem}     

\usepackage[table]{xcolor}      
\definecolor{highlight}{HTML}{D55E00}  
\usepackage[normalem]{ulem}    
\usepackage{xspace}
\usepackage{pifont}             

\usepackage[round,authoryear]{natbib}

\usepackage[colorlinks=true,
            linkcolor=blue,
            citecolor=blue,
            urlcolor=blue,
            bookmarksopen=true]{hyperref}
\usepackage[capitalise,noabbrev]{cleveref}

\newcommand{\system}{\textsc{SparseRL-Sync}\xspace}
\newcommand{\eg}{e.g.,\xspace}

\title{\system: Lossless Weight Synchronization with $\sim$\,$100\times$ Less Communication}

\author{%
\makebox[\textwidth][c]{\small %
  Lucas Hu$^{*}$,\ %
  Ranchi Zhao$^{*}$,\ %
  Isaac Zhu,\ %
  Zach Zhang,\ %
  Hscos Zhang,\ %
  Hugh Yin,\ %
  Jason Zhao$^{\dagger}$%
}\\[2pt]
\makebox[\textwidth][c]{\small \href{https://www.scitix.ai/}{Scitix}}\\[2pt]
\makebox[\textwidth][c]{\footnotesize $^{*}$Equal contribution.\quad $^{\dagger}$Corresponding author.}
}

\date{\today}
\begin{document}
\maketitle
\begingroup
\renewcommand\thefootnote{}
\footnotetext{Codes will be released at \url{https://github.com/scitix/helix}.}
\endgroup
\vspace{-3.0em}
\begin{abstract}
    In large-scale reinforcement learning (RL) systems with decoupled Trainer--Rollout execution, the Trainer must regularly synchronize policy weights to the Rollout side to limit policy staleness. When inter-node bandwidth is abundant, such synchronization is usually only a small fraction of end-to-end cost. As model size grows, however, the communication demand rises rapidly. In bandwidth-constrained or network-variable deployments---for example, cross-datacenter or cross-cluster settings, heterogeneous resource pools, and online RL---weight synchronization can become a dominant bottleneck for throughput and tail latency.

    We observe that, in mainstream large-model RL training, the locations where parameters actually change are highly sparse at the element level (often $99\%+$ sparsity). Building on this observation, we propose and implement \system, which replaces full-weight transfers with a \emph{lossless} sparse update payload (indices and values) that can be \emph{exactly} reconstructed on the inference side, thereby preserving $100\%$ fidelity. Under a simplified cost model, sparse synchronization reduces the per-update communication volume from $S$ to approximately $S/X$; with $99\%$ sparsity ($X\!\approx\!100$), this yields about a $100\times$ reduction in transmitted data. Combined with appropriate bucketing, \system{} also reduces launch and control-plane overhead, significantly improving scalability and end-to-end efficiency in bandwidth-limited and highly asynchronous RL settings. 
\end{abstract}


\section{Introduction}
\label{sec:introduction}

\begin{figure}[H]
  \centering
  \begin{subfigure}[t]{0.52\linewidth}
    \includegraphics[width=\linewidth]{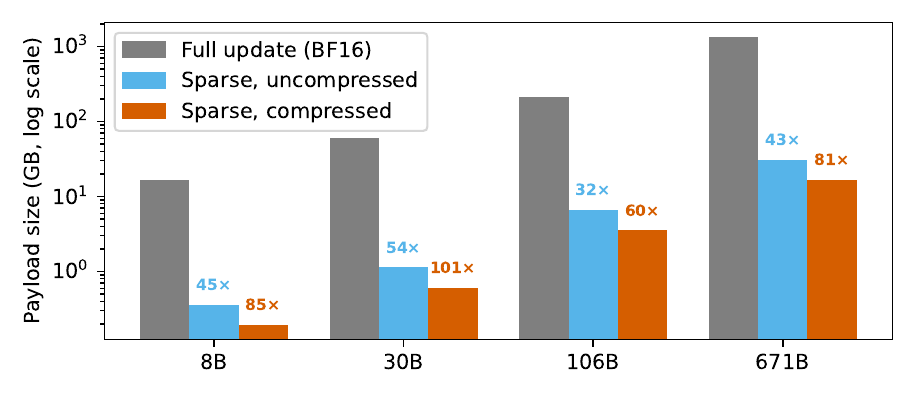}
    \caption{Per-synchronization payload across model scales: full update (BF16) vs.\ sparse $(I,V)$ uncompressed vs.\ sparse $(I,V)$ compressed. Sparse synchronization reduces the transfer by $32\times$--$54\times$ raw and $\approx\!60\times$--$101\times$ after lossless compression (\Cref{sec:method:compress}).}
    \label{fig:intro:payload}
  \end{subfigure}\hfill
  \begin{subfigure}[t]{0.45\linewidth}
    \includegraphics[width=\linewidth]{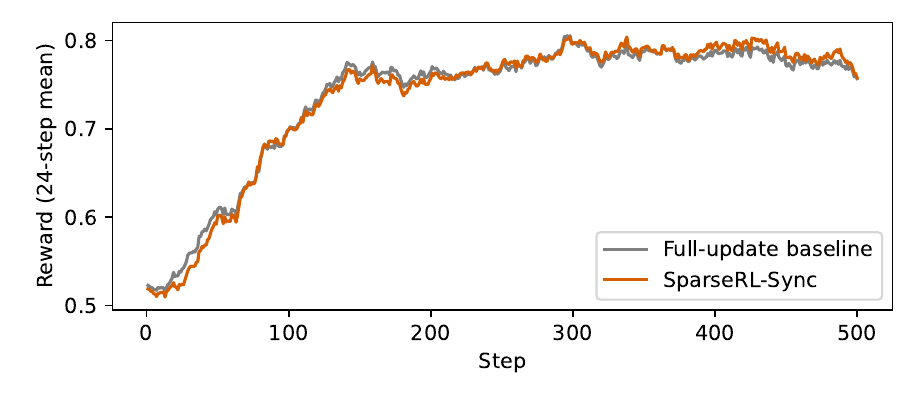}
    \caption{Reward curves of the full-update baseline vs.\ \system{} on Qwen3-30B-A3B-Instruct-2507 over 500 training steps. The two curves are nearly indistinguishable, confirming lossless fidelity.}
    \label{fig:intro:reward}
  \end{subfigure}
  \caption{\textbf{Core result at a glance.} \system{} reduces the Trainer-to-Rollout weight-synchronization payload by $32\times$--$54\times$ raw and up to $\approx\!100\times$ after lossless compression across model scales (left) while preserving training dynamics bit-exactly (right).}
  \label{fig:intro:overview}
\end{figure}

\textcolor{black}{Large-model RL training systems typically decouple the \emph{Trainer} (training) from the \emph{Rollout} (inference) component. The Trainer computes losses and updates model parameters from collected trajectories, while the Rollout uses the current policy to generate new trajectories. To limit policy staleness and preserve training stability, updated parameters must be synchronized regularly from Trainer to Rollout.} \textcolor{black}{This interaction can be understood from two complementary perspectives: the data exchanged between Trainer and Rollout, and the way the two components are deployed. We describe these two aspects in turn.}

\paragraph{Two interaction flows.}
\textcolor{black}{Trainer and Rollout interact through two distinct data flows.}
\begin{itemize}
  \item \textbf{Sample data flow:} Rollout performs inference and sampling, producing trajectories, tokens, and rewards that are returned to the Trainer. \textcolor{black}{This is the primary training path and is typically less bandwidth-intensive than weight synchronization.}
  \item \textbf{Weights data flow:} After one or more optimization steps, the Trainer synchronizes updated policy weights to Rollout. The payload size scales directly with model size, is typically much larger than the sample flow, and quickly becomes the bottleneck in bandwidth-constrained settings.
\end{itemize}

\paragraph{Two placement strategies.}
\textcolor{black}{Trainer and Rollout are typically deployed in one of two ways.}
\begin{itemize}
  \item \textbf{Time-sharing:} \textcolor{black}{The two components share the same set of GPUs and alternate between training and inference through time multiplexing. In this case, weight synchronization is typically limited to intra-process switching or intra-node transfer (\eg CUDA IPC).}
  \item \textbf{Space-sharing:} \textcolor{black}{GPUs are partitioned between Trainer and Rollout so that training and inference can proceed concurrently. Weight updates must then be distributed across multiple GPUs or nodes, typically via collective communication (e.g., NCCL), making the system much more sensitive to network bandwidth and topology.}
\end{itemize}

\textcolor{black}{In large-model RL, Rollout's \texttt{generate()} is often the end-to-end throughput bottleneck and exhibits pronounced tail latency. To mitigate this bottleneck and improve overall GPU utilization, many systems adopt asynchronous RL (Async-RL), which is commonly implemented through space-sharing or fully disaggregated deployment to further decouple Trainer and Rollout. This design makes parameter synchronization a central scalability bottleneck.}

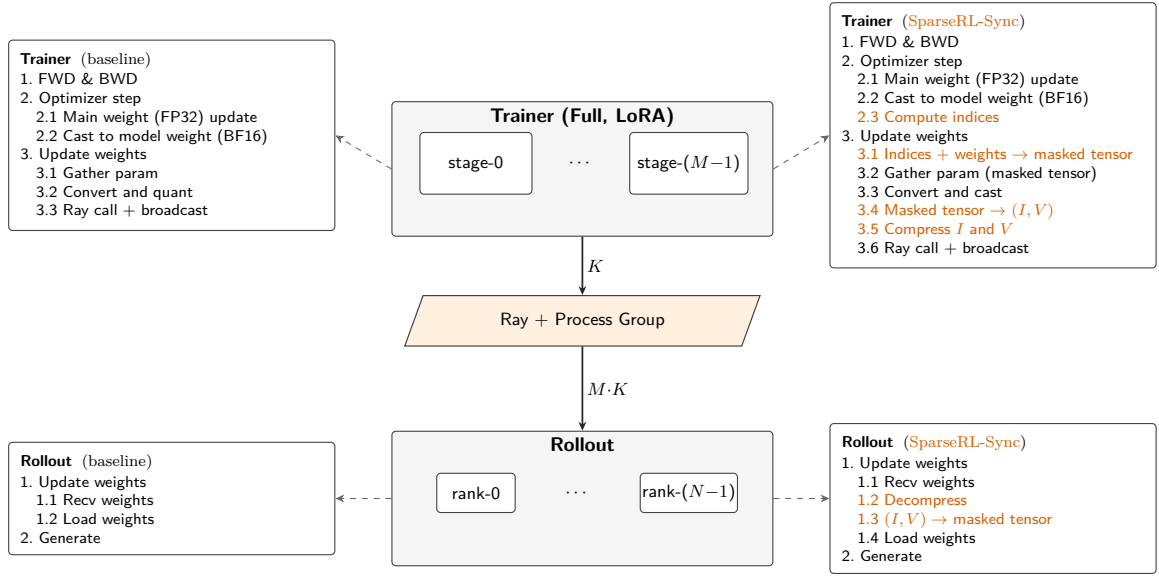
\begin{figure}[t]
  \centering
  \resizebox{0.96\linewidth}{!}{
%
%
\begin{tikzpicture}[
  font=\sffamily\footnotesize,
  >={Stealth[length=1.6mm,width=1.4mm]},
  every node/.append style={align=center, inner sep=2pt},
  blk/.style={draw=black!75, rounded corners=2pt, line width=0.55pt,
              inner sep=4pt},
  outer/.style={blk, fill=black!4, minimum height=24mm, minimum width=68mm},
  stage/.style={blk, fill=white, minimum width=20mm, minimum height=11mm},
  pg/.style={draw=black!75, line width=0.6pt, fill=orange!12,
             trapezium, trapezium left angle=70, trapezium right angle=110,
             minimum width=34mm, minimum height=9mm, inner sep=3pt},
  rank/.style={blk, fill=white, minimum width=14mm, minimum height=7mm,
               inner sep=1pt},
  detail/.style={blk, fill=white, text width=54mm, align=left,
                 inner sep=6pt, font=\sffamily\scriptsize},
  dethead/.style={font=\sffamily\small\bfseries},
  arr/.style={->, line width=0.55pt, draw=black!85},
  darr/.style={dashed, ->, line width=0.55pt, draw=black!60},
  hi/.style={text=highlight},
]
  \node[outer] (trainer) {};
  \node[anchor=north, yshift=-1pt, font=\sffamily\small\bfseries]
    at (trainer.north) {Trainer (Full, LoRA)};
  \node[stage] (s0) at ($(trainer.center)+(-19mm,1mm)$) {stage-0};
  \node[stage] (sm) at ($(trainer.center)+( 19mm,1mm)$) {stage-($M{-}1$)};
  \node at ($(s0.east)!0.5!(sm.west)$) {$\cdots$};

  \node[pg] (pg) at ($(trainer.south)+(0,-15mm)$) {Ray + Process Group};

  \node[outer, below=15mm of pg] (rollout) {};
  \node[anchor=north, yshift=-1pt, font=\sffamily\small\bfseries]
    at (rollout.north) {Rollout};

  \node[rank] (r0) at ($(rollout.center)+(-19mm,1mm)$) {rank-0};
  \node[rank] (rn) at ($(rollout.center)+( 19mm,1mm)$) {rank-($N{-}1$)};
  \node at ($(r0.east)!0.5!(rn.west)$) {$\cdots$};

  \draw[arr, line width=0.9pt] (trainer.south) --
        node[right,pos=0.5] {$K$} (pg.north);
  \draw[arr, line width=0.9pt] (pg.south) --
        node[right,pos=0.5] {$M{\cdot}K$} (rollout.north);

  \node[detail, anchor=east] (dTrainerL) at ($(trainer.west)+(-10mm,6mm)$)
    {%
      \textbf{Trainer \;\textnormal{\scriptsize(baseline)}}\\[1pt]
      1.\ FWD \& BWD\\
      2.\ Optimizer step\\
      \quad 2.1\ Main weight (FP32) update\\
      \quad 2.2\ Cast to model weight (BF16)\\
      3.\ Update weights\\
      \quad 3.1\ Gather param\\
      \quad 3.2\ Convert and quant\\
      \quad 3.3\ Ray call + broadcast\\
    };

  \node[detail, anchor=east] (dRollL) at ($(rollout.west)+(-10mm,0)$)
    {%
      \textbf{Rollout \;\textnormal{\scriptsize(baseline)}}\\[1pt]
      1.\ Update weights\\
      \quad 1.1\ Recv weights\\
      \quad 1.2\ Load weights\\
      2.\ Generate\\
    };

  \node[detail, anchor=west] (dTrainerR) at ($(trainer.east)+(10mm,6mm)$)
    {%
      \textbf{Trainer \;\textnormal{\scriptsize(\textcolor{highlight}{SparseRL-Sync})}}\\[1pt]
      1.\ FWD \& BWD\\
      2.\ Optimizer step\\
      \quad 2.1\ Main weight (FP32) update\\
      \quad 2.2\ Cast to model weight (BF16)\\
      \quad {\color{highlight}2.3\ Compute indices}\\
      3.\ Update weights\\
      \quad {\color{highlight}3.1\ Indices $+$ weights $\to$ masked tensor}\\
      \quad 3.2\ Gather param (masked tensor)\\
      \quad 3.3\ Convert and cast\\
      \quad {\color{highlight}3.4\ Masked tensor $\to$ $(I,V)$}\\
      \quad {\color{highlight}3.5\ Compress $I$ and $V$}\\
      \quad 3.6\ Ray call + broadcast\\
    };

  \node[detail, anchor=west] (dRollR) at ($(rollout.east)+(10mm,0)$)
    {%
      \textbf{Rollout \;\textnormal{\scriptsize(\textcolor{highlight}{SparseRL-Sync})}}\\[1pt]
      1.\ Update weights\\
      \quad 1.1\ Recv weights\\
      \quad {\color{highlight}1.2\ Decompress}\\
      \quad {\color{highlight}1.3\ $(I,V) \to$ masked tensor}\\
      \quad 1.4\ Load weights\\
      2.\ Generate\\
    };

  \draw[darr] (trainer.west) -- (dTrainerL.east);
  \draw[darr] (rollout.west) -- (dRollL.east);
  \draw[darr] (trainer.east) -- (dTrainerR.west);
  \draw[darr] (rollout.east) -- (dRollR.west);
\end{tikzpicture}}
  \caption{Trainer--Rollout weight-synchronization workflow. \textbf{Left column:} the full-update baseline pipeline used by open-source RL frameworks such as \texttt{slime}. \textbf{Right column:} the \system{} pipeline, with newly inserted steps highlighted in vermillion. The center column shows the physical topology shared by both: $M$ Trainer stages (PP size)\,$\to$\,Ray\,+\,process group\,$\to$\,$N$ Rollout ranks. Each Trainer stage contributes $K$ buckets to the broadcast, and the process group forwards the aggregated $M{\cdot}K$ buckets to every rank. \system{} reduces each bucket's payload via sparse $(I,V)$ encoding without altering the collective topology.}
  \label{fig:workflow}
\end{figure}

\textcolor{black}{This bottleneck is rooted in the way current systems perform Trainer-to-Rollout weight synchronization.} \Cref{fig:workflow} (left column) illustrates the parameter-update pipeline in the open-source RL framework \textcolor{black}{\texttt{slime}~\citep{slime_framework}}. After a number of optimizer steps, the Trainer (i)~gathers parameters along the TP and EP dimensions, (ii)~performs format conversion and any necessary quantization, and (iii)~broadcasts from Trainer to Rollout ranks.

\begin{figure}[t]
  \includegraphics[width=\linewidth]{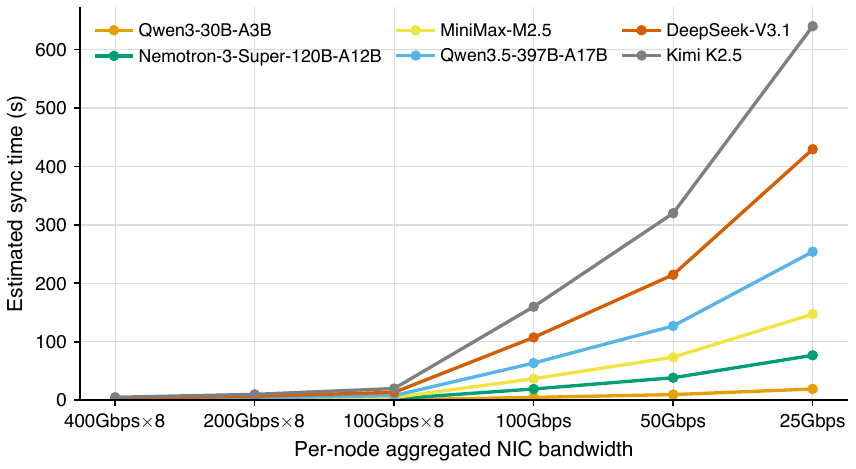}
  \caption{Estimated wall-clock cost of a single full-weight (BF16) parameter update for representative open models under different per-node aggregated NIC bandwidths, including Qwen3-30B-A3B (30B), Nemotron-3-Super-120B-A12B (120B), MiniMax-M2.5 (230B), Qwen3.5-397B-A17B (397B), DeepSeek-V3.1 (671B), and Kimi K2.5 (1TB). As model size increases and available bandwidth decreases, the synchronization cost rises sharply.}
  \label{fig:NICs:cat}
\end{figure}

\textcolor{black}{The cost of this design becomes increasingly pronounced as model size grows and available bandwidth decreases.} \Cref{fig:NICs:cat} shows the estimated cost of performing a single parameter update for several representative models under different communication bandwidths. With high update frequency and large payloads, the cost grows almost linearly with model size and inversely with bandwidth, making weight synchronization a first-class scalability concern.

\paragraph{Existing engineering remedies.}
To make parameter synchronization work in practice, current systems combine a series of engineering techniques:
\begin{itemize}
  \item \textbf{Bucketing and pipelining:} split the weight tensor into multiple buckets to reduce peak memory and shorten tail latency.
  \item \textbf{Communication--computation overlap:} hide synchronization behind the compute path, at the cost of additional thread/stream/scheduling complexity.
  \item \textbf{Eliminating redundant transfers:} avoid retransmitting identical weight versions, eliminate extra copies, and avoid amplification caused by merging-before-sending.
  \item \textbf{Bandwidth-aware scheduling:} when broadcasting to many Rollout nodes, schedule the distribution order and concurrency so that slow nodes do not stall the whole system.
\end{itemize}

\textcolor{black}{
These techniques mitigate synchronization overhead in important ways, but they do not change the communication object itself: the synchronized payload remains the full-weight tensor. As model size grows and deployment becomes increasingly bandwidth-constrained, this design choice remains a fundamental scalability bottleneck. Our work revisits this assumption. Rather than further optimizing the full-update synchronization path, we exploit the element-level sparsity of the BF16 weight delta sent from Trainer to Rollout and redesign synchronization around a lossless sparse update representation.}

\paragraph{Contributions.}
\begin{itemize}
  \item We \emph{extend the empirical scope} of the sparsity observation beyond prior RLVR results on a fixed model to a range of mainstream RL settings, including GRPO, DAPO, GSPO, asynchronous RL, and agentic RL, across dense and MoE models ranging from 8B to 671B parameters, and under BF16, FP16, and FP8 synchronization (\Cref{sec:sparsity:empirical,sec:experiments}).

  \item We design \system, a \emph{lossless} sparse-synchronization mechanism for Trainer-to-Rollout weight updates, and develop a precise cost model for the sparse $(I,V)$ payload together with lossless encoding schemes for both indices and values that lift the raw compression ratio of $32\times$--$54\times$ to a compressed ratio of $\approx\!60\times$--$101\times$ across model scales from 8B to 671B (\Cref{sec:method,sec:method:compress}).

  \item We validate \system{} end-to-end in two complementary studies: a correctness study over $500$ training steps confirms \emph{bit-exact} equivalence to the full-update baseline and an indistinguishable reward trajectory, and a performance study across two bandwidth regimes shows that sparse synchronization substantially reduces Trainer-to-Rollout broadcast time in both high- and low-bandwidth environments (\Cref{sec:experiments}).
\end{itemize}

\begin{table}[t]
  \centering
  \small
  \setlength{\tabcolsep}{4pt}
  \renewcommand{\arraystretch}{1.15}
  \caption{Comparison of sparse-synchronization support and reported performance across recent RL training systems. ``Validated size'' is the largest model scale for which weight synchronization has been publicly reported; ``Perf.''\ is the reported synchronization latency at that scale. ``Validated in Agentic-RL'' indicates whether sparse synchronization has been validated specifically under agentic RL workloads. Performance figures for third-party systems are taken from publicly available technical reports, blog posts, or official documentation and are reproduced here for reference only; figures marked ``?'' were not publicly disclosed at the time of writing.}
  \label{tab:sparse_sync_compare}
  \begin{tabularx}{\linewidth}{>{\raggedright\arraybackslash}X@{\hspace{3pt}}cccc}
    \toprule
    \textbf{System} & \textbf{Sparse sync} & \textbf{Validated in} & \textbf{Validated size} & \textbf{Perf.} \\
                    &                      & \textbf{Agentic-RL} & & \\
    \midrule
    Slime / Miles          & \ding{55} & \ding{55} & $\sim$\,TB     & $7$--$40$\,s \\
    MoonshotAI / Checkpoint Engine & \ding{55} & \ding{55} & $\sim$\,TB     & $\le 20$\,s   \\
    AWex                   & \ding{55} & \ding{55} & $\sim$\,TB     & $\le 10$\,s   \\
    Composer2              & \ding{51} & \ding{51} & $\sim$\,TB     & ?             \\
    Pulse                  & \ding{51} & \ding{55} & $\sim 7$\,B    & ?             \\
    \rowcolor{gray!10}
    \system\ (ours)        & \ding{51} & \ding{51} & $\sim$\,TB     & $\approx 1$\,s \\
    \bottomrule
  \end{tabularx}
\end{table}

\section{Sparsity Analysis}
\label{sec:sparsity}
The starting point of our system design is an empirical regularity: in mainstream large-model RL training, the BF16 model-weight delta sent to Rollout at every synchronization point is almost entirely sparse at the element level. \Cref{sec:sparsity:empirical} documents this regularity across five RL settings; \Cref{sec:sparsity:theory} summarizes the explanation given by the Three-Gate Theory of \citet{zhu2025pathnottaken} for why this happens. \textcolor{black}{The theory itself is borrowed; our contributions are the empirical extension to MoE / GRPO~\citep{shao2024grpo} / DAPO~\citep{yu2025dapo} / GSPO~\citep{zheng2025gspo} / Async-RL / Agentic-RL settings and its system-level exploitation in \Cref{sec:method}.}

\subsection{Empirical Observations}
\label{sec:sparsity:empirical}

\paragraph{Setup.}
\textcolor{black}{We instrument Qwen3-30B-A3B-Instruct-2507 during RL fine-tuning and capture snapshots of the BF16 model weights immediately before and after each offline weight-update event. From these snapshots, we compute two statistics, both per tensor and aggregated across the model: (i) the fraction of elements whose BF16 value changes (\emph{element-level update ratio}, with sparsity defined as its complement), and (ii) the fraction of tensors for which no element changes (\emph{tensor-level inactive ratio}). We use BF16 for both training and inference, and otherwise follow the default configuration of \texttt{slime}. We repeat this measurement under five RL settings: GRPO, DAPO, GSPO, an asynchronous RL variant (Async-RL), and an agentic RL variant (Agentic-RL).}

\begin{figure}[!htbp]
  \centering
  \includegraphics[width=0.86\linewidth]{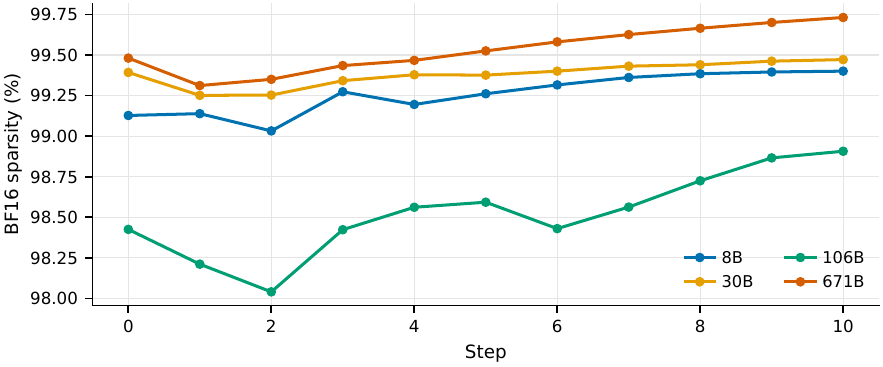}
  \caption{BF16 element-level sparsity across the four model scales (8B, 30B, 106B, 671B) over synchronization steps. All models exhibit high sparsity ($\geq\!98\%$) from the first step, and sparsity tends to increase over training. The 671B model reaches the highest observed sparsity, consistent with larger pretrained weights having more mass in the sub-threshold regime that gets absorbed by the BF16 cast.}
  \label{fig:model_scale_sparsity}
\end{figure}

\paragraph{Models under study.}
Where measurements are taken on a single model, we use Qwen3-30B-A3B-Instruct-2507. For the cross-scale study in \Cref{fig:model_scale_sparsity} and throughout the paper, we evaluate four models spanning dense and MoE architectures from $8$\,B to $671$\,B parameters: \textbf{Qwen3-8B-Base} (abbreviated \textbf{8B}), \textbf{Qwen3-30B-A3B-Instruct-2507} (\textbf{30B}), \textbf{GLM-4.5-Air-Base} (\textbf{106B}), and \textbf{DeepSeek-V3.1-Base} (\textbf{671B}). For brevity, we refer to each model by its parameter-count abbreviation in subsequent figures, tables, and prose. We next summarize the main empirical observations that emerge from these measurements.


\paragraph{Pervasive element-level sparsity.}
\begin{table}[!htbp]
  \centering
  \small
  \caption{\textcolor{black}{Sparsity summary on 30B across five RL settings. BF16 sparsity and inactive tensors are measured at the synchronized working-precision weights; FP32 change ratio is measured on the Trainer-side main weights.}}
  \label{tab:sparsity_summary}
  \setlength{\tabcolsep}{4pt}
  \renewcommand{\arraystretch}{1.12}
  \begin{tabular}{lrrrr}
    \toprule
    \textbf{Setting} & \textbf{BF16 sparse} & \textbf{BF16 changed} & \textbf{FP32 changed} & \textbf{Inactive tensors} \\
    \midrule
    DAPO       & $99.4003\%$ & $0.5997\%$ & $99.4704\%$ & $4.9732\%$ \\
    GRPO       & $99.3859\%$ & $0.6141\%$ & $99.4595\%$ & $5.5216\%$ \\
    GSPO       & $99.3958\%$ & $0.6042\%$ & $99.4653\%$ & $5.9673\%$ \\
    Async-RL   & $99.4006\%$ & $0.5994\%$ & $99.4354\%$ & $5.3852\%$ \\
    Agentic-RL & $99.3030\%$ & $0.6970\%$ & $99.6845\%$ & $5.4193\%$ \\
    \bottomrule
  \end{tabular}
\end{table}

\Cref{tab:sparsity_summary} reports the per-step element-level sparsity averaged across an entire training run. Despite the diversity of objective functions, all five settings consistently reach $99.30\%$--$99.40\%$ sparsity, with the last-step sparsity slightly higher than the run-wide mean. The implication is direct: at any synchronization point, fewer than $1\%$ of BF16 weight elements actually need to be transmitted to Rollout.

\paragraph{BF16 vs.\ FP32: a precision-gated sparsity gap.}
\begin{figure}[!htbp]
  \centering
  \begin{subfigure}[t]{0.48\linewidth}
    \centering
    \includegraphics[width=\linewidth]{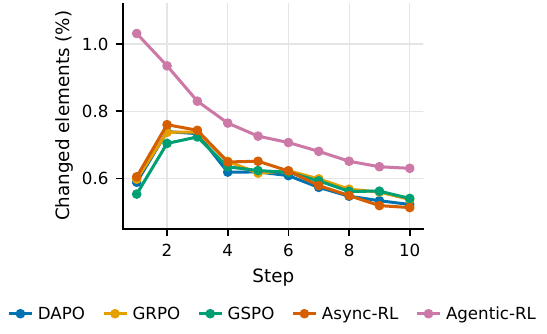}
    \caption{BF16 model parameters.}
    \label{fig:origin_precision_gap:bf16}
  \end{subfigure}\hfill
  \begin{subfigure}[t]{0.48\linewidth}
    \centering
    \includegraphics[width=\linewidth]{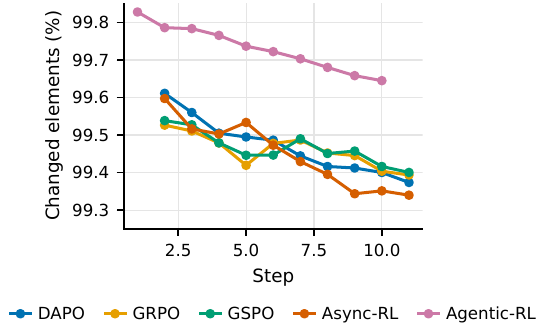}
    \caption{FP32 master weights.}
    \label{fig:origin_precision_gap:fp32}
  \end{subfigure}
  \caption{Precision-gated sparsity gap over synchronization steps. (a)~BF16 model parameters synchronized to Rollout have sub-$1\%$ changed-element density; (b)~FP32 master weights on the Trainer side remain near-dense throughout. Both panels share the same algorithm legend, shown below each subfigure.}
  \label{fig:origin_precision_gap}
\end{figure}

\Cref{tab:sparsity_summary,fig:origin_precision_gap} show that as training progresses, \textcolor{black}{the element-level change ratio of \emph{BF16} weights} steadily decreases, dropping from roughly $0.75\%$ in the early steps to below $0.56\%$ by step~8. By contrast, the Trainer-side FP32 \emph{main} weights remain near-dense throughout, with the element-level change ratio consistently above $99.4\%$. This precision-dependent gap is consistent across all five RL settings.

This behavior is not accidental. Most updates produce micro-changes that fall below the BF16 quantization threshold; they are therefore \emph{absorbed} during the FP32-to-BF16 cast and become invisible after rounding. In other words, the sparsity emerges at the precision-conversion stage rather than in the underlying FP32 update itself. We provide a quantitative explanation of this precision-gated effect in \Cref{sec:sparsity:theory}. This precision-gated gap is the foundation of the lossless sparse-sync mechanism described in \Cref{sec:method}.


\paragraph{Precision controls visible sparsity.}
\begin{figure}[!htbp]
  \centering
  \includegraphics[width=\linewidth]{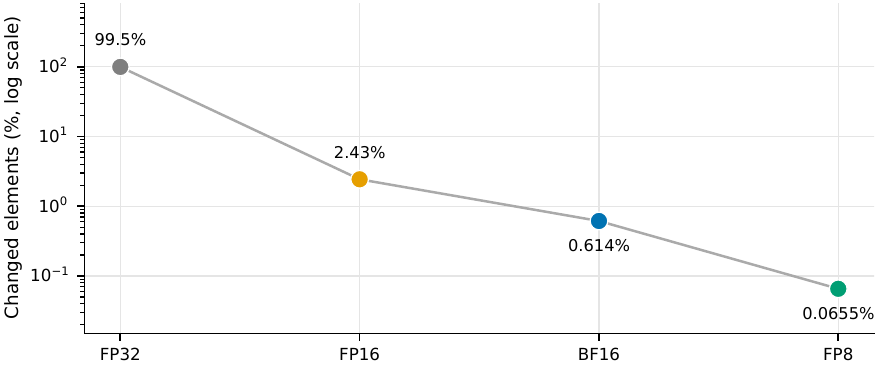}
  \caption{Element-level changed-element density under different synchronization precisions, on a log scale so that the FP16 / BF16 / FP8 differences remain visible. Values shown are measured on Qwen3-30B-A3B over a GRPO run.}
  \label{fig:origin_precision_sparsity_compare}
\end{figure}

\Cref{fig:origin_precision_sparsity_compare} compares the apparent sparsity induced by different numerical formats. FP32 main weights have only $0.54\%$ sparsity in the GRPO run, meaning that almost every FP32 element changes after an optimizer update. After casting to BF16, the same synchronization boundary exposes only $0.6141\%$ changed elements. FP16, despite having higher numerical precision than BF16 overall, has a \emph{finer} mantissa ($10$ bits vs.\ $7$ bits) and therefore absorbs fewer micro-updates at the cast boundary, exposing $2.4308\%$ changed elements and $97.5692\%$ sparsity---more than BF16. FP8 has a coarser visibility threshold and exposes only $0.0655\%$ changed elements, yielding $99.9345\%$ sparsity. These measurements show that the sparse-update opportunity is precision dependent: it is weak in FP32, strong in FP16, stronger in BF16, and strongest in FP8.

\paragraph{Element-level vs.\ tensor-level sparsity.}
\begin{figure}[!htbp]
  \centering
  \includegraphics[width=0.86\linewidth]{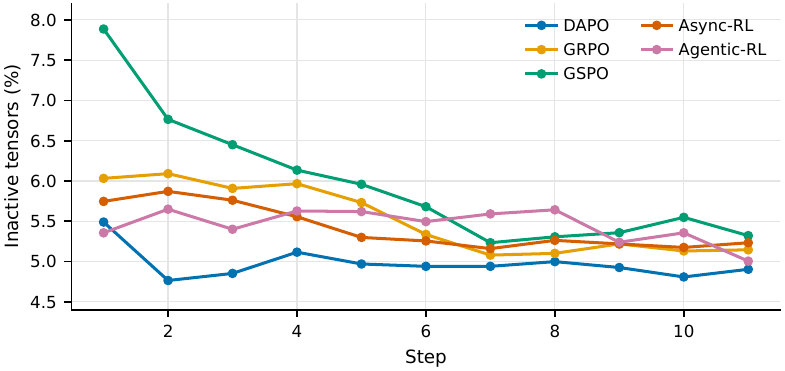}
  \caption{Tensor-level inactive ratio over synchronization steps. Only about $5\%$--$6\%$ of parameter tensors have no changed elements, so the observed sparsity is primarily within tensors.}
  \label{fig:origin_tensor_inactive}
\end{figure}

\Cref{tab:sparsity_summary,fig:origin_tensor_inactive} switch to the parameter-tensor level and report the fraction of tensors for which \emph{no} element changed. Across the five settings, the inactive-tensor ratio is only about $5\%$--$6\%$, so more than $94\%$ of tensors still contain at least one changed element at each synchronization point. The sparsity is therefore structural \emph{within} tensors, not between them. This rules out the naive optimization ``skip tensors that did not change'', and shows that any practical sparse-sync mechanism must operate \emph{at the element level}.

This conclusion also holds for the expert weights of MoE models, where one might suspect that routing concentrates updates on a small subset of experts. On 30B, we tracked the update status of each expert tensor over $10$ consecutive synchronization steps: only $14$ expert tensors ($0.46\%$) were never updated in any step, while $3{,}010$ expert tensors ($97.98\%$) were updated in \emph{every} one of the $10$ steps. Even among experts---the structural component most plausibly amenable to coarse-grained skipping---tensor-level inactivity is a vanishing minority. The sparse-update opportunity is genuinely at the element level.

\paragraph{Sparsity across model scales.}

\Cref{fig:model_scale_sparsity} extends the sparsity observation to all four model scales (8B, 30B, 106B, 671B). All four models exhibit $\geq\!98\%$ BF16 sparsity from the very first synchronization step, confirming that high update sparsity is not a property of any particular model scale or architecture. The 671B model climbs to $\geq\!99.5\%$ sparsity within a handful of steps and stays there, while the 106B MoE model shows somewhat lower sparsity ($\sim\!98\%$--$99\%$), likely reflecting architectural differences in weight magnitude distributions. Across all scales, sparsity tends to increase over training.

\paragraph{Temporal locality of update indices.}
\begin{figure}[!htbp]
  \centering
  \includegraphics[width=\linewidth]{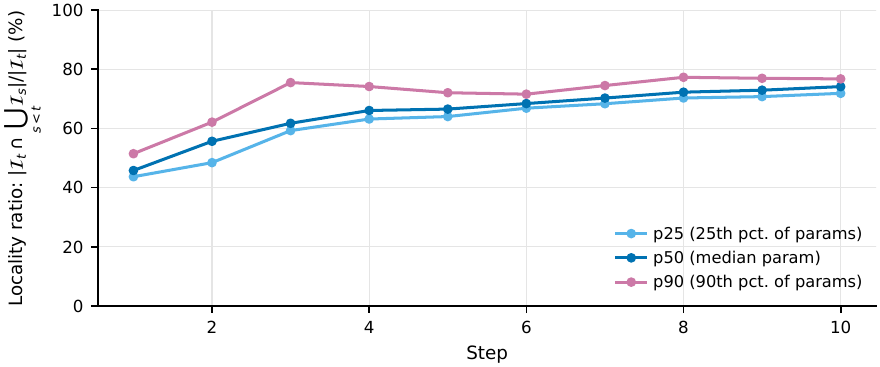}
  \caption{Temporal locality of update indices on 30B (GRPO). For each sync step $t$ and each parameter tensor, we compute the locality ratio $|\mathcal{I}_t \cap \bigcup_{s<t}\mathcal{I}_s|\,/\,|\mathcal{I}_t|$---the fraction of the current changed indices that have appeared in any prior step. The three curves show the 25th, 50th (median), and 90th percentiles of this ratio across all parameter tensors. All three rise monotonically from $\sim\!45\%$--$52\%$ at step~1 to $\sim\!72\%$--$77\%$ by step~10, indicating that the majority of updated weight positions recur from earlier synchronization events.}
  \label{fig:index_locality}
\end{figure}

Beyond element-level sparsity, the changed indices exhibit strong \emph{temporal locality}: the set of weight elements updated at step $t$ overlaps substantially with the union of indices from all prior steps. \Cref{fig:index_locality} quantifies this on 30B (GRPO). For each step $t$ and each parameter tensor, the locality ratio $|\mathcal{I}_t \cap \bigcup_{s<t}\mathcal{I}_s|\,/\,|\mathcal{I}_t|$ measures what fraction of the current changed indices reappear from history. The p25/p50/p90 quantiles across parameter tensors all rise monotonically from $\sim\!45\%$--$52\%$ at step~1 to $\sim\!72\%$--$77\%$ by step~10. The tight spread between quantiles confirms that this locality is a consistent property across parameter tensors, not driven by a small subset of outlier parameters. This temporal concentration indicates that a persistent ``hot'' subspace of weight elements accounts for the majority of updates across the training run, and is a promising direction for future delta-index compression schemes.

\paragraph{Takeaways.}
We summarize the empirical regularities exploited by \system:
\begin{enumerate}
  \item \textbf{Pervasive element-level sparsity.} Across all RL settings we tested, the BF16 update is $\sim\!99.4\%$ sparse on average and tends to become sparser over training.
  \item \textbf{Precision-gated gap.} The same updates are near-dense in FP32 main weights but highly sparse in BF16 model weights, meaning that the sparsity arises at the FP32-to-BF16 cast.
  \item \textbf{Precision-dependent visibility.} FP16 exposes more changed elements than BF16 (finer mantissa absorbs fewer micro-updates), while FP8 exposes fewer; the sparsity is a property of the synchronization precision, not only of the optimizer step.
  \item \textbf{Structural within-tensor sparsity.} Almost every tensor is touched on every step, so naive tensor-skipping is insufficient; element-level indexing is required.
  \item \textbf{Cross-scale universality.} High BF16 update sparsity ($\geq\!98\%$) holds across model scales from 8B to 671B and across dense and MoE architectures.
  \item \textbf{Temporal locality.} For each parameter tensor, the fraction of its current changed indices that appeared in any prior step rises monotonically from $\sim\!45\%$ at step~1 to $\sim\!72\%$ by step~10 (median across tensors), indicating a persistent ``hot'' subspace of frequently updated elements.
\end{enumerate}

\subsection{Mechanistic Explanation: The Three-Gate Theory}
\label{sec:sparsity:theory}

\citet{zhu2025pathnottaken} provide a principled explanation for why RL fine-tuning of pretrained large language models yields apparently sparse parameter updates. We adopt their Three-Gate theory here as the explanatory framework for our observations, not as a new theoretical contribution of our own, but as a compact mechanistic account of why the BF16 weight delta synchronized from Trainer to Rollout is highly sparse across the RL settings we study. This framework directly motivates both the design of \system{} (\Cref{sec:method}) and the ablation experiments used to test it (\Cref{sec:experiments}).


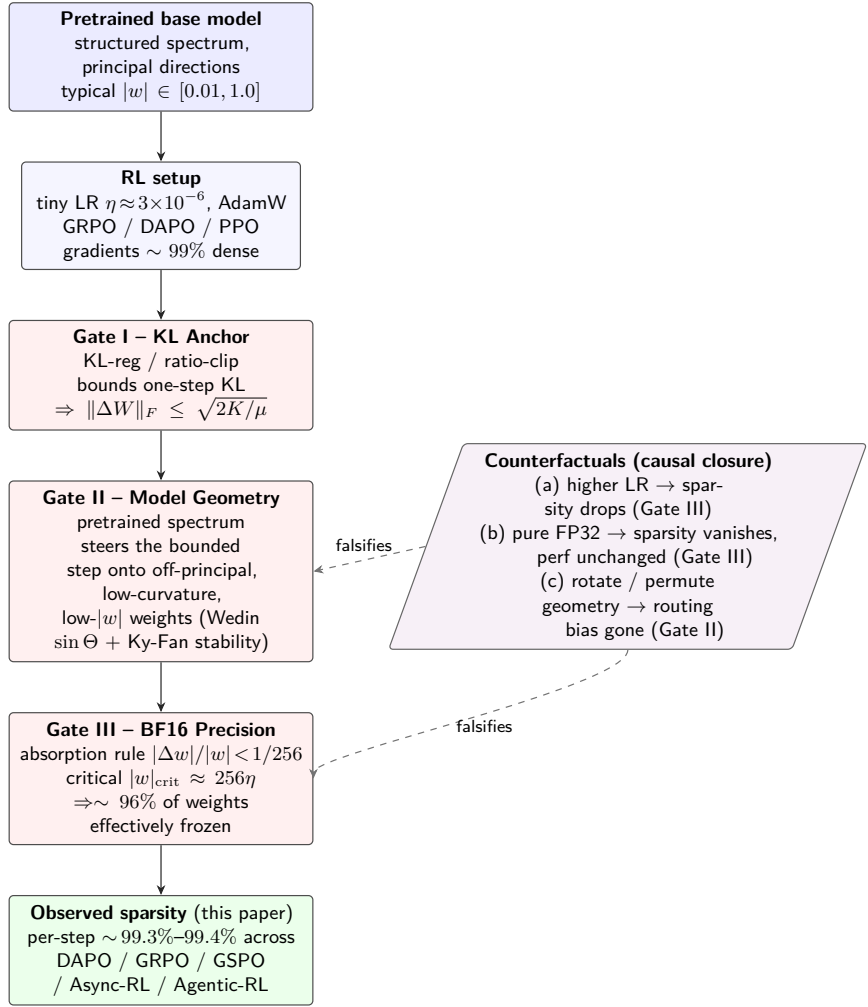
\begin{figure}[t]
  \centering
  \resizebox{0.72\linewidth}{!}{
%
\begin{tikzpicture}[
  font=\sffamily\footnotesize,
  >={Stealth[length=1.6mm,width=1.4mm]},
  every node/.append style={align=center, inner sep=4pt, line width=0.55pt},
  pre/.style ={draw=black!70, fill=blue!8,   rounded corners=2pt,
               text width=46mm},
  setup/.style={draw=black!70, fill=blue!4,  rounded corners=2pt,
                text width=42mm},
  gate/.style={draw=black!70, fill=red!6,    rounded corners=2pt,
               text width=46mm},
  obs/.style={draw=black!70, fill=green!8,   rounded corners=2pt,
              text width=46mm},
  cf/.style ={draw=black!60, fill=violet!6,  rounded corners=2pt,
              text width=50mm,
              shape=trapezium, trapezium left angle=70,
              trapezium right angle=110, align=center},
  arr/.style={->, line width=0.55pt, draw=black!85},
  darr/.style={dashed, ->, line width=0.55pt, draw=black!55},
]
  \node[pre] (pre) {%
    \textbf{Pretrained base model}\\
    structured spectrum, principal directions\\
    typical $|w|\!\in\![0.01,1.0]$};

  \node[setup, below=8mm of pre] (setup) {%
    \textbf{RL setup}\\
    tiny LR $\eta\!\approx\!3{\times}10^{-6}$, AdamW\\
    GRPO / DAPO / PPO\\
    gradients $\sim\!99\%$ dense};

  \node[gate, below=8mm of setup] (g1) {%
    \textbf{Gate I -- KL Anchor}\\
    KL-reg / ratio-clip bounds one-step KL\\
    $\Rightarrow \|\Delta W\|_F \le \sqrt{2K/\mu}$};

  \node[gate, below=8mm of g1] (g2) {%
    \textbf{Gate II -- Model Geometry}\\
    pretrained spectrum steers the bounded\\
    step onto off-principal, low-curvature,\\
    low-$|w|$ weights (Wedin $\sin\Theta$ +
    Ky-Fan stability)};

  \node[gate, below=8mm of g2] (g3) {%
    \textbf{Gate III -- BF16 Precision}\\
    absorption rule $|\Delta w|/|w|\!<\!1/256$\\
    critical $|w|_{\mathrm{crit}}\!\approx\!256\eta$\\
    $\Rightarrow \sim\!96\%$ of weights effectively frozen};

  \node[obs, below=8mm of g3] (obs) {%
    \textbf{Observed sparsity} (this paper)\\
    per-step $\sim\!99.3\%$--$99.4\%$ across\\
    DAPO / GRPO / GSPO / Async-RL / Agentic-RL};

  \node[cf, right=18mm of g2, yshift=4mm] (cfs) {%
    \textbf{Counterfactuals (causal closure)}\\
    (a) higher LR $\to$ sparsity drops (Gate III)\\
    (b) pure FP32 $\to$ sparsity vanishes,\\
    \quad\ \ perf unchanged (Gate III)\\
    (c) rotate / permute geometry $\to$ routing\\
    \quad\ \ bias gone (Gate II)};

  \draw[arr] (pre)   -- (setup);
  \draw[arr] (setup) -- (g1);
  \draw[arr] (g1)    -- (g2);
  \draw[arr] (g2)    -- (g3);
  \draw[arr] (g3)    -- (obs);

  \draw[darr] (cfs.west)  -- node[above,pos=0.55,font=\sffamily\scriptsize]
                              {falsifies} (g2.east);
  \draw[darr] (cfs.south) .. controls +(0,-6mm) and +(0,4mm) ..
                              node[right,pos=0.55,font=\sffamily\scriptsize]
                              {falsifies} (g3.east);
\end{tikzpicture}}
  \caption{Three-Gate Theory of \citet{zhu2025pathnottaken}, reproduced here as the explanatory framework for our sparsity observations. The pretrained base model and the RL optimizer (top) jointly set a small-step, KL-bounded regime; three successive gates---Gate~I (KL anchor) bounds step magnitude, Gate~II (model geometry) routes the bounded update onto off-principal, low-curvature coordinates, and Gate~III (BF16 precision) suppresses sub-threshold updates at the FP32$\!\to\!$BF16 cast---together account for the $\sim\!99\%$ element-level sparsity we measure. The box on the right lists counterfactual probes from \citet{zhu2025pathnottaken} that isolate the contribution of individual gates.}
  \label{fig:theory}
\end{figure}

As summarized in \Cref{fig:theory}, the apparent sparsity of RL weight updates arises from three coupled effects: a KL-constrained small-step regime, geometry-induced routing away from dominant pretrained directions, and BF16 rounding that suppresses many resulting micro-updates. Together, these effects explain why the underlying FP32 update can remain nearly dense while the BF16 weight delta seen by Rollout becomes highly sparse. The conclusion we draw---and that underlies \system{}---is that the $\sim\!99\%$ sparsity we observe is not an artifact of any single RL algorithm or model family, but a structural consequence of the small-step, low-precision regime in which contemporary large-model RL typically operates.
\section{Method: \system}
\label{sec:method}

Building on the empirical pervasiveness of element-level update sparsity (\Cref{sec:sparsity:empirical}), we now present \system, a \emph{lossless} sparse-synchronization mechanism that replaces full-weight synchronization with $(\text{indices},\text{values})$ messages reconstructible bit-for-bit on the Rollout side. Throughout this section we use \emph{master weights} ($W^{\mathrm{main}}$) for the high-precision parameter copy maintained by the optimizer (e.g., FP32 in standard mixed-precision training) and \emph{model parameters} ($W$) for the working-precision copy used in forward and backward passes (e.g., BF16).

\subsection{Design Goals and Overview}
\label{sec:method:overview}

\paragraph{Design goals.}
\system{} is engineered to satisfy four properties:
\begin{description}
  \item[\textbf{G1.\ Lossless fidelity.}] Rollout receives \emph{exactly} the model weights the Trainer has, so the policy used for sampling is identical to the one being trained.
  \item[\textbf{G2.\ Drop-in integration.}] The Trainer-side change is local to the optimizer-step boundary; the Rollout-side change is local to the weight-loader.
  \item[\textbf{G3.\ Bandwidth reduction $\propto$ sparsity.}] The on-wire payload should scale as $\Theta(|\mathcal{I}|)$ in the number of changed elements.
  \item[\textbf{G4.\ Universality.}] One mechanism covers dense / MoE models, full fine-tuning / LoRA.
\end{description}

\paragraph{Overall workflow.}
Recall \Cref{fig:workflow} from \Cref{sec:introduction}: the right column depicts the \system{} pipeline and contrasts it with the full-update baseline on the left. Four additional steps (highlighted in vermillion) are inserted into the existing pipeline:
\begin{enumerate}
  \item \textit{Compute indices.} After the master weights are cast and copied into the model parameters, the Trainer compares the pre- and post-copy values elementwise to obtain the set of changed indices.
  \item \textit{Materialize a change-masked tensor.} Using these indices, the Trainer constructs a parameter-shaped tensor initialized with a sentinel value (NaN), then writes the current model-parameter values at the changed positions. Unchanged positions remain as NaN and carry no information.
  \item \textit{Convert to $(I,V)$.} Before transmission, the sparse tensor is converted into an $(\text{indices},\text{values})$ payload (optionally losslessly encoded; \Cref{sec:method:compress}).
  \item \textit{Reconstruct on Rollout.} Each Rollout rank applies the received $(I,V)$ updates as an in-place scatter into its local weight buffer.
\end{enumerate}
The control plane is unchanged: synchronization events are still triggered via Ray remote calls, and the underlying parameter broadcast uses the same PyTorch process groups (NCCL) as before; the only structural difference is that each bucket's on-wire size is reduced in proportion to the element-level update sparsity.

\subsection{Algorithms}
\label{sec:method:algorithms}

The mechanism is implemented by three cooperating procedures: (1)~the Trainer collects changed indices alongside its optimizer step (\Cref{alg:sparserl-trainer-indices}); (2)~a WeightUpdater module packs the indices and values into a self-describing message (\Cref{alg:sparserl-weight-updater-pack}); (3)~each Rollout rank applies the message in place (\Cref{alg:sparserl-rollout-apply}).

\begin{algorithm}[t]
\caption{Trainer: optimizer step with index tracking}
\label{alg:sparserl-trainer-indices}
\begin{algorithmic}[1]
\Require Model parameters $W$ (e.g., BF16); master weights $W^{\mathrm{main}}$ (e.g., FP32); steps $1{:}T$.
\Ensure  Updated $W$ and the cumulative changed-index set $\mathcal{I}_T$.
\State $\mathcal{I}_0 \gets \emptyset$
\For{$t \gets 1$ \textbf{to} $T$}
  \State \Call{OptimizerStep}{$t$}                            \Comment{updates $W^{\mathrm{main}}$}
  \State $W_{\text{prev}} \gets \mathrm{detach\_clone}(W)$
  \State \Call{CastAndCopy}{$W \leftarrow \mathrm{round}_{\mathrm{BF16}}(W^{\mathrm{main}})$}
  \State $I_t \gets \{\, i \mid W^{(i)} \neq W_{\text{prev}}^{(i)} \,\}$       \Comment{precision filter: working-precision cast absorbs sub-threshold updates}
  \State $\mathcal{I}_t \gets \mathcal{I}_{t-1} \cup I_t$
\EndFor
\State \Return $(W, \mathcal{I}_T)$
\end{algorithmic}
\end{algorithm}

\Cref{alg:sparserl-trainer-indices} accumulates element-level deltas across $T$ optimizer steps. The cumulative set $\mathcal{I}_T$ is the set of model-parameter elements that need to be communicated at the next synchronization point, and is by construction \emph{precision-aware}: indices that vanish at the precision-reducing cast from $W^{\mathrm{main}}$ to $W$ are never inserted. Note that $\mathcal{I}_T$ is a conservative \emph{superset} of the true post-$T$ delta against the last synchronization snapshot: an element that changed at some intermediate step but whose final value equals the pre-sync value still appears in $\mathcal{I}_T$. This overapproximation is harmless for correctness (Algorithm~\ref{alg:sparserl-weight-updater-pack} transmits the current values $W[\mathcal{I}_T]$, so Rollout receives a bit-exact copy of $W$ regardless of redundant indices) and only costs a bounded amount of extra bandwidth proportional to the superset size. The snapshot $W_{\text{prev}}$ is freed immediately after $I_t$ is computed, so this step adds no persistent memory overhead.

\begin{algorithm}[t]
\caption{WeightUpdater: pack and broadcast sparse updates}
\label{alg:sparserl-weight-updater-pack}
\begin{algorithmic}[1]
\Require Named tensors $\{(\textit{name}, \textit{param}, \mathcal{I})\}$ with changed-index set $\mathcal{I}$.
\Ensure  Sparse update message delivered to Rollout.
\State $\textit{meta} \gets [\,];\quad I\_list \gets [\,];\quad V\_list \gets [\,]$
\ForAll{$(\textit{name}, \textit{param}, \mathcal{I})$}
  \State $M \gets \Call{Materialize}{\textit{param},\, \mathcal{I}}$  \Comment{param-shaped; NaN at unchanged positions, \textit{param}$^{(i)}$ at $i\!\in\!\mathcal{I}$}
  \State $U \gets \Call{ConvertForBroadcast}{M}$                      \Comment{gather complete parameter tensor, fix layout and dtype}
  \State $\mathcal{V} \gets U[\mathcal{I}]$                            \Comment{$\mathcal{I}\!:\!\texttt{int32}$, $\mathcal{V}\!:\!\texttt{BF16}$}
  \State $(\mathcal{I}, \mathcal{V}) \gets \Call{OptionalEncode}{\mathcal{I}, \mathcal{V}}$  \Comment{see \Cref{sec:method:compress}}
  \State \textbf{append} $(\textit{name}, \mathrm{dtype}(U), \mathrm{shape}(U))$ \textbf{to} $\textit{meta}$
  \State \textbf{append} $\mathcal{I}$ \textbf{to} $I\_list$;\quad \textbf{append} $\mathcal{V}$ \textbf{to} $V\_list$
\EndFor
\State \Call{SendToRollout}{$\textit{meta}, I\_list, V\_list$}
\end{algorithmic}
\end{algorithm}

\Cref{alg:sparserl-weight-updater-pack} packs every changed parameter into a self-describing payload $(\textit{meta}, I\_list, V\_list)$. Indices are kept in \texttt{int32} (so the encoding is correct for any tensor shape that fits in $2^{31}$ flattened elements, i.e.\ all parameter tensors in current open large models including DeepSeek-V3.1-Base). \textsc{ConvertForBroadcast} assembles the complete parameter tensor from its distributed shards and applies any required layout and dtype conversion; values $\mathcal{V}$ are then extracted from this fully-assembled tensor, so the receiver does not need any knowledge of the Trainer's parallelism layout. The optional encode step, \Cref{sec:method:compress}, is where additional bandwidth gains compound on top of the raw $(I,V)$ cost.

\begin{algorithm}[t]
\caption{Rollout: receive and apply sparse updates}
\label{alg:sparserl-rollout-apply}
\begin{algorithmic}[1]
\Require Local weights $W$.
\Ensure  Updated local weights $\widetilde W$.
\State $(\textit{meta}, I\_list, V\_list) \gets \Call{RecvFromUpdater}{\,}$
\For{$k \gets 1$ \textbf{to} $|\textit{meta}|$}
  \State $(\textit{name}, \textit{dtype}, \textit{shape}) \gets \textit{meta}[k]$
  \State $\mathcal{I} \gets I\_list[k];\quad \mathcal{V} \gets V\_list[k]$
  \State $(\mathcal{I}, \mathcal{V}) \gets \Call{OptionalDecode}{\mathcal{I}, \mathcal{V}}$
  \State $W[\textit{name}][\mathcal{I}] \gets \mathcal{V}$       \Comment{in-place sparse scatter}
\EndFor
\State \Return $\widetilde W$
\end{algorithmic}
\end{algorithm}

\Cref{alg:sparserl-rollout-apply} mirrors \Cref{alg:sparserl-weight-updater-pack} on the Rollout side. The optional decode is the exact inverse of the optional encode, and the sparse scatter $W[\textit{name}][\mathcal{I}] \gets \mathcal{V}$ is implemented as a fused kernel. Because the values transmitted are the bit-exact post-cast model-parameter values, after this step the local Rollout weights are bit-identical to the Trainer's model parameters---this is the property G1 (lossless fidelity).

\subsection{Cost Model and Lossless Compression}
\label{sec:method:compress}

\paragraph{Raw $(I,V)$ cost model.}
Let $S = N \cdot b_v$ be the size in bytes of a full working-precision weight broadcast ($N$ elements at $b_v$ bytes each), and let $\rho \in [0,1]$ be the element-level update density (so $1-\rho$ is the sparsity). Encoding indices as fixed-width \texttt{int32} integers ($b_i = 4$\,B) alongside BF16 values ($b_v = 2$\,B), the sparse payload has size
\begin{equation}
  S_{\mathrm{sparse}}(\rho) \;=\; \rho N \,(b_v + b_i) \;+\; S_{\mathrm{meta}},
  \label{eq:cost-model}
\end{equation}
where $S_{\mathrm{meta}}$ is the constant per-tensor metadata overhead. Ignoring $S_{\mathrm{meta}}$, the raw compression ratio is
\begin{equation}
  X(\rho) \;=\; \frac{S}{S_{\mathrm{sparse}}(\rho)} \;\approx\; \frac{b_v}{\rho\,(b_v + b_i)} \;=\; \frac{1}{3\rho}.
  \label{eq:compression}
\end{equation}

\paragraph{Lossless compression of the $(I,V)$ payload.}
The raw formula of \cref{eq:compression} treats the indices and values as unstructured binary blobs. We apply two complementary lossless transforms before transmission and their exact inverses on the receiver, preserving G1 (lossless fidelity) exactly.

\textit{Index delta encoding.} Because the changed indices $\mathcal{I}$ are sorted, we store the first-differences $\Delta\mathcal{I}$ (prepending a zero) instead of absolute values. Empirically, the maximum inter-index gap for typical linear-weight tensors is well below $2^{15}$ (measured maxima are in the low thousands), so the deltas fit in \texttt{int16} ($b_i = 2$\,B), halving the index stream. Embedding-table tensors, whose indices can span up to $\sim\!10^8$ positions, retain \texttt{int32} ($b_i = 4$\,B).

\textit{Value entropy coding.} The BF16 value stream $\mathcal{V}$ is also highly compressible: changed weight values cluster near their previous magnitudes, yielding a distribution that standard lossless entropy coders exploit well. Empirically, entropy coding reduces the value stream to $\alpha \in [0.60, 0.70]$ of its original size.

Combining both passes, the compressed payload size becomes
\begin{equation}
  S_{\mathrm{compressed}}(\rho) \;=\; \rho N \,(b_i + \alpha\, b_v),
  \label{eq:cost-compressed}
\end{equation}
where $b_i = 2$\,B (delta-encoded \texttt{int16} for linear weights) or $4$\,B (embedding tables), and the combined compression ratio is
\begin{equation}
  X_{\mathrm{compressed}}(\rho) \;=\; \frac{b_v}{\rho\,(b_i + \alpha\, b_v)}.
  \label{eq:compression-full}
\end{equation}

\paragraph{A concrete example reaching $\approx\!100\times$.}
The compression ratio in \cref{eq:compression-full} depends only on $\rho$, $b_v$, $b_i$, and $\alpha$---not on the absolute model size. We illustrate with 30B, which reaches a mean element-level sparsity of $99.38\%$ over GRPO fine-tuning (\Cref{fig:model_scale_sparsity}), corresponding to an update density of $\rho = 0.62\%$. Using delta-encoded \texttt{int16} indices ($b_i = 2$\,B) and value entropy coding at $\alpha = 0.60$, the combined ratio is
\begin{equation*}
  X_{\mathrm{compressed}}(0.0062) \;=\; \frac{2}{0.0062\,(2 + 0.60\times 2)} \;\approx\; 100\times.
\end{equation*}
Equivalently, each changed element occupies $b_i + \alpha b_v = 3.2$\,B on the wire versus $b_v = 2$\,B for every element in the full-update baseline, so the per-parameter sparse cost is $1.6\rho$ of the full-update cost. Across the four model scales (8B--671B) reported in \Cref{fig:intro:payload}, $\rho$ remains below $1.1\%$, yielding raw $(I,V)$ ratios of $32\times$--$54\times$ and compressed ratios of $60\times$--$101\times$.

\subsection{Integration}
\label{sec:method:integration}

\paragraph{Trainer.}
The hook sits at the optimizer-step epilogue, between the high-precision optimizer update to the master weights and the cast-and-copy step that materializes them into model parameters (\textsc{CastAndCopy} in \Cref{alg:sparserl-trainer-indices}). Immediately after this step writes the new model-parameter values, we diff the new and previous parameter buffers to collect $\mathcal{I}_t$. This change is local to the optimizer-step boundary and is independent of the parallelism backend.

\paragraph{WeightUpdater.}
Triggered by the RL framework via a Ray remote call at each synchronization event. The WeightUpdater iterates over named parameters and routes each one based on its update density: parameters with high sparsity follow the $(I,V)$ path of \Cref{alg:sparserl-weight-updater-pack}; parameters that change on nearly every element each step are transmitted via a full-weight copy. For example, LoRA adapter weights---which are small by design and exhibit near-$100\%$ update density---take the full-copy path, while frozen base-model parameters exhibiting $\geq\!99\%$ sparsity take the sparse path. The routing is per-parameter and transparent to the training loop.

\paragraph{Rollout.}
The hook patches the weight-loader's \texttt{tensor.copy\_()} call. Instead of copying a full parameter tensor, the patched copy accepts an $(I,V)$ payload, unpacks it into a NaN-masked buffer, and applies an in-place sparse scatter (\Cref{alg:sparserl-rollout-apply}). The same patch applies to any inference framework that loads weights through a parameter copy step.


\section{Experiments}
\label{sec:experiments}

Our evaluation answers two questions:
\begin{enumerate}
  \item \textbf{Correctness} (\Cref{sec:exp:correctness}): does sparse $(I,V)$ reconstruction preserve the RL trajectory bit-exactly, and is the reward curve indistinguishable from a full-update baseline?
  \item \textbf{Communication savings} (\Cref{sec:exp:latency}): how much does \system{} reduce the on-wire payload and transmission time across model scales and bandwidth regimes?
\end{enumerate}

\subsection{Setup}
\label{sec:exp:setup}

\paragraph{Framework and models.}
All runs use \textbf{Helix}~\citep{scitix_helix}, our in-house RL framework, with Megatron-LM for the Trainer and SGLang for the Rollout. Integration follows \Cref{sec:method:integration}: a Trainer-side hook at the optimizer cast boundary (\Cref{alg:sparserl-trainer-indices}) and a Rollout-side patch at the weight-loader boundary (\Cref{alg:sparserl-rollout-apply}). We use 30B for the correctness study, and report communication savings in the TB-scale regime where bandwidth is the dominant cost: 106B (measured) and 671B (projected).

\paragraph{Hardware.}
Each node is a single-socket $8\times$H100-SXM5 server with $4$ RDMA NICs (one NIC per two GPUs).

\paragraph{Bandwidth regimes.}
Point-to-point GPU--GPU benchmarks from one 8-GPU node to $15$ peers define two inter-node regimes used throughout:
\begin{itemize}
  \item \textbf{IB on} (RDMA active): per-GPU $34.99$\,GB/s mean ($30.3$--$37.9$); per-node $\approx\!280$\,GB/s.
  \item \textbf{IB off} (TCP fallback): per-GPU $2.84$\,GB/s mean ($0.91$--$5.30$); per-node $\approx\!22.7$\,GB/s.
\end{itemize}
Intra-node (NVLink) is $\approx\!319$\,GB/s per GPU. The two inter-node regimes bracket the deployment envelope from well-provisioned RDMA clusters to cross-cluster / TCP-only settings, where sparse synchronization matters most.

\subsection{Correctness Validation}
\label{sec:exp:correctness}

\paragraph{Bit-exact reconstruction.}
At each synchronization event the Rollout first applies the full-weight update and snapshots the result, then restores the previous state and applies the sparse $(I,V)$ update. Comparing the two copies tensor by tensor, all layers match bit-for-bit on every synchronization event across the full run.

\paragraph{Reward validation.}
\Cref{fig:intro:reward,tab:reward_validate} compare rollout-level rewards of a full-update baseline and a \system{} run on 30B over 500 steps. The curves are visually indistinguishable: mean-reward diff $-8\times10^{-6}$, MAE $0.0186$, per-rollout Pearson correlation $0.9749$.

\begin{table}[!htbp]
  \centering
  \small
  \caption{Reward-level validation on 30B over $500$ steps.}
  \label{tab:reward_validate}
  \setlength{\tabcolsep}{4.5pt}
  \begin{tabular}{lrrrrr}
    \toprule
    \textbf{Metric} & \textbf{Full-update mean} & \textbf{\system{} mean} & \textbf{Diff.} & \textbf{MAE} & \textbf{Corr.} \\
    \midrule
    Reward & $0.738095$ & $0.738087$ & $-8\times10^{-6}$ & $0.0186$ & $0.9749$ \\
    \bottomrule
  \end{tabular}
\end{table}

These results confirm that \system{} does not measurably perturb the sampled reward trajectory.

\subsection{Communication Savings Across Model Scales}
\label{sec:exp:latency}

\paragraph{Setup.} We measure 106B on $128$ H100 GPUs in separated (disaggregated) mode, split evenly as $64$ Trainer + $64$ Rollout. A separated 671B deployment would require more than $128$ GPUs, so we instead \emph{project} its broadcast time by applying the 106B effective bandwidth to the 671B full-update payload ($1\,342$\,GB) and sparse payload ($\approx\!31.0$\,GB at $\rho\!\approx\!0.77\%$). The projection is conservative: larger-scale collective broadcasts typically achieve equal or better utilization than the 106B baseline. All \system{} numbers in this subsection use the raw $(I,V)$ path; the additional lossless index/value compression of \Cref{sec:method:compress} would shrink the \system{} payloads further, which we leave to future measurements.

\begin{figure}[htbp]
  \centering
  \includegraphics[width=0.88\linewidth]{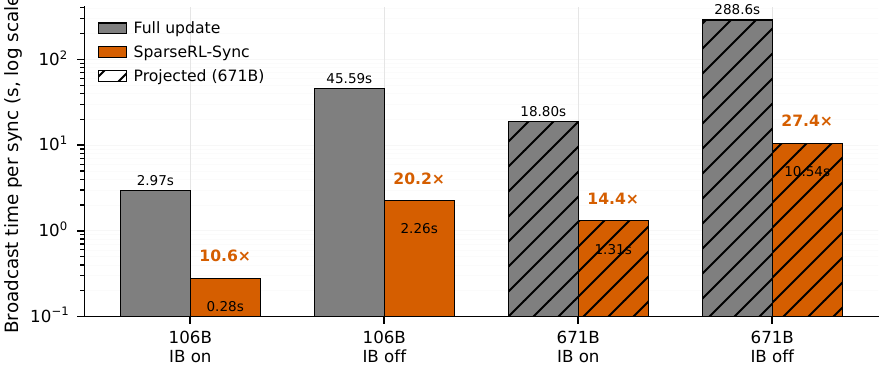}
  \caption{Per-synchronization broadcast time under the two bandwidth regimes of \Cref{sec:exp:setup}. 106B is measured on $128$ H100 GPUs in separated mode ($64$ Trainer + $64$ Rollout); 671B is projected from the 106B effective bandwidth (hatched bars). Numbers on top of each \system{} bar are speedups over the corresponding full-update baseline. Note the log-scale $y$-axis.}
  \label{fig:comm_savings}
\end{figure}

\paragraph{Findings.} Three observations stand out.
\textbf{(i)~IB-off becomes usable.} Without RDMA, a full 106B broadcast takes $45.6$\,s; \system{} brings it down to $2.26$\,s---matching the single-digit budget that the full-update path only achieves with RDMA. The same pattern projects to 671B: nearly $5$\,minutes collapses to $\approx\!10.5$\,s.
\textbf{(ii)~Speedup grows as bandwidth shrinks.} We observe $10.6\times$--$14.4\times$ under IB on and $20.2\times$--$27.4\times$ under IB off, consistent with the cost model of \Cref{sec:method:compress}: once the payload is an order of magnitude smaller, its transmission time depends far less on the bandwidth gap between regimes.
\textbf{(iii)~Sync cadence improves by an order of magnitude.} The 106B IB-off drop from $45.59$\,s to $2.26$\,s moves Trainer$\,\to\,$Rollout synchronization from ``many-second'' to ``low-second,'' the threshold at which Async-RL deployments no longer need to hide synchronization behind rollout tail latency.
\FloatBarrier

%
%

\section{Related Work}
\label{sec:related}

\system{} is most closely related to three lines of prior work. The first is the algorithmic literature on RL fine-tuning of large language models, which defines the optimization regime in which our system operates. The second is the systems literature on weight synchronization and gradient/weight compression for distributed training, which addresses similar bandwidth bottlenecks and provides technical building blocks that we re-use. The third is the recent literature on the parameter-space dynamics of RLVR, which provides the mechanistic basis for our sparsity observations.

\subsection{RL Fine-Tuning of Large Language Models}
\label{sec:related:rl}
Reinforcement learning from human feedback~\citep{ouyang2022rlhf} and Proximal Policy Optimization (PPO)~\citep{schulman2017ppo} laid the foundation for RL-based LLM post-training. In long-horizon reasoning settings, however, PPO's reliance on a learned value function can introduce substantial credit-assignment and optimization overhead~\citep{kazemnejad2024vineppo}, motivating critic-free alternatives such as Group Relative Policy Optimization (GRPO)~\citep{shao2024grpo}, which estimates advantages from multiple rollouts per prompt.

A central challenge in this line of work is the stability of importance-ratio and advantage-based updates. DAPO~\citep{yu2025dapo} introduces decoupled clipping and dynamic sampling for large-scale LLM RL; SAPO~\citep{gao2025sapo} replaces hard clipping with a smooth, adaptive gate; and GSPO~\citep{zheng2025gspo} moves importance-ratio control from the token level to the sequence level. Related work further studies off-policy or sample-polarity effects, including tapered off-policy REINFORCE~\citep{leroux2025topr}, asymmetric importance-sampling correction~\citep{wang2025aspo}, adaptive advantage shaping~\citep{tang2025a3po}, and negative-enhanced GRPO~\citep{nan2025ngrpo}.

\system{} is \emph{algorithm-agnostic}: it operates on the post-cast BF16 weight delta and does not modify the loss, the importance ratio, or the optimizer. We therefore treat these algorithms as benchmark settings (\Cref{sec:sparsity:empirical}) rather than as competitors.

\subsection{Weight Synchronization and Compression in Distributed Training}
\label{sec:related:systems}

The communication cost of Trainer-to-Rollout weight synchronization is determined by model size, synchronization frequency, and the network conditions between Trainer and Rollout. As RL deployments move toward larger models and increasingly disaggregated resource pools, this path has become a first-class systems concern. Recent systems have therefore optimized the weight-update path in several complementary ways.

\paragraph{Engineering optimized full-weight systems.}
A first line of systems work focuses on making \emph{full}-weight transfer practical through engineering optimizations:
\begin{itemize}
  \item \textit{Slime/Miles}~\citep{slime_framework} provides an open-source RL post-training framework that connects high-performance training with rollout/inference backends. Its full-update Trainer-to-Rollout weight-update pipeline serves as the baseline path illustrated in \Cref{fig:workflow} (left column).
  \item \textit{Kimi checkpoint engine}~\citep{moonshot_checkpoint_engine} reports efficient in-place weight updates for a 1T-parameter Kimi-K2 model across thousands of GPUs in approximately 20 seconds.
  \item \textit{AWex}~\citep{antgroup_awex} is a high-performance RL training--inference weight-synchronization framework designed to enable second-level parameter updates from training to inference, with support for heterogeneous deployment modes and transfer optimizations.
\end{itemize}
These systems substantially reduce synchronization latency, but they leave the underlying communication object unchanged: the payload is still a full BF16 or quantized weight tensor, so the transmitted data volume scales with model size rather than with the number of elements that actually changed.

\paragraph{Sparse and delta-based systems.}
A second line of work reduces the communicated payload by sending deltas or sparse updates:
\begin{itemize}
  \item \textit{Composer2}~\citep{cursor_composer2} describes a geographically distributed RL infrastructure in which inference clusters reconstruct weights from a shared delta chain over commodity cloud storage. The public report describes the high-level delta-chain design, but does not fully disclose the sparse encoding details or provide a public implementation of the synchronization layer.
  \item \textit{PULSE} by Miahi and Belilovsky~\citep{miahi2026weightsparsity} performs the closest analysis of weight-update sparsity in distributed RL and proposes a lossless sparse update path that transmits the indices and values of modified parameters. Their public evaluation focuses on bandwidth-constrained decentralized RL settings, whereas \system{} targets MoE models, TB-scale synchronization, and integration with production training stacks.
\end{itemize}

\system{} extends this line of work along three axes: (a) coverage of MoE architectures up to 671B parameters, (b) support for both Megatron-LM and FSDP-style training stacks, and (c) support for both full fine-tuning and LoRA. We also release an open-source implementation integrated with \texttt{slime} and SGLang.

\paragraph{Communication-efficient distributed SGD.}
A long line of work studies the more general problem of compressing \emph{gradients} in data-parallel SGD, including 1-bit SGD~\citep{seide20141bit}, TernGrad~\citep{wen2017terngrad}, top-$k$ sparsification~\citep{aji2017sparse}, Deep Gradient Compression~\citep{lin2018dgc}, and PowerSGD~\citep{vogels2019powersgd}. Our setting differs in two crucial ways. First, we communicate rounded model weights rather than gradients, so we do not require error feedback to compensate for dropped gradient information. Second, we require lossless reconstruction of the inference-visible weights, which rules out lossy compressors such as sign-based, top-$k$, or low-rank approximations. We therefore reuse only the encoding tools from this literature, such as compact index encodings and entropy coding, while avoiding lossy approximation of the communicated values.

\subsection{Parameter-Space Dynamics of RLVR}
\label{sec:related:rlvr-dynamics}

A complementary line of work asks not \emph{how} to compress RL updates, but \emph{why} they appear sparse in the first place. \citet{zhu2025pathnottaken} provide a parameter-level account of RLVR training and propose the Three-Gate Theory summarized in \Cref{sec:sparsity:theory}: a KL-constrained small-step regime, pretrained model geometry, and low-precision rounding jointly make the post-cast weight delta appear sparse. They also contrast this regime with supervised fine-tuning, which follows different parameter-space dynamics. Our paper takes this explanation as an external theoretical account and operationalizes it as a systems-level design invariant in the RL settings we study: because the BF16 synchronization delta is consistently sparse, we can build a lossless sparse-update infrastructure around it.

\paragraph{Asynchronous and heterogeneous RL.}
Decoupling Trainer and Rollout is itself an active systems direction. AReaL~\citep{areal2024async} explores fully asynchronous RL, in which generation is decoupled from training to improve GPU utilization. Composer2~\citep{cursor_composer2} demonstrates geographically distributed rollout infrastructure across multiple clusters. ROLL~\citep{roll2024heterogeneous} provides a large-scale RL framework for LLM training over large GPU resources and heterogeneous training scenarios. These systems make efficient Trainer--Rollout synchronization increasingly important. \system{} is complementary to them: rather than changing the RL algorithm or deployment topology, it reduces the synchronized payload itself.
\section{Conclusion}
\label{sec:conclusion}
\label{sec:conclusion}
We presented \system{}, a lossless sparse-synchronization mechanism for the Trainer-to-Rollout weight path in large-model reinforcement learning. We observe that the post-cast model-weight delta synchronized to Rollout is highly sparse at the element level across mainstream RL settings, meaning that full-weight transfer contains substantial redundant communication. Based on this observation, \system{} replaces full-weight broadcasts with sparse $(I,V)$ messages that transmit only changed indices and their updated values while preserving bit-exact equivalence to full-update synchronization. Our implementation integrates with Megatron-LM, FSDP, \texttt{slime}, and SGLang, supports dense and MoE models as well as full fine-tuning and LoRA, and reduces weight-synchronization volume by $32\times$--$54\times$ at the raw $(I,V)$ level and $60\times$--$101\times$ with lossless index/value compression across model scales from 8B to 671B.

\bibliographystyle{plainnat}
\bibliography{refs}

\end{document}